\pdfoutput=1

\documentclass[11pt]{article}

\usepackage{acl}

\usepackage{times}
\usepackage{latexsym}

\usepackage[T1]{fontenc}

\usepackage[utf8]{inputenc}

\usepackage{microtype}

\usepackage{CJKutf8}
\newcommand\DEL[1]{}
\usepackage{graphicx}
\usepackage{multirow}
\usepackage[T2A,T1]{fontenc}
\usepackage[russian,english]{babel}

\title{Addressing Segmentation Ambiguity in Neural Linguistic Steganography}

\author{Jumon Nozaki
  \hspace{25mm} Yugo Murawaki \\
    Graduate School of Informatics, Kyoto University \\
    Yoshida-honmachi, Sakyo-ku, Kyoto, 606-8501, Japan \\
  \texttt{nozaki@sap.ist.i.kyoto-u.ac.jp}
  \hspace{10mm}\texttt{murawaki@i.kyoto-u.ac.jp}
}

\begin{document}
\maketitle
\begin{abstract}
Previous studies on neural linguistic steganography, except \citet{ueoka-etal-2021-frustratingly}, overlook the fact that the sender must detokenize cover texts to avoid arousing the eavesdropper's suspicion.
In this paper, we demonstrate that segmentation ambiguity indeed causes occasional decoding failures at the receiver's side.
With the near-ubiquity of subwords, this problem now affects any language.
We propose simple tricks to overcome this problem, which are even applicable to languages without explicit word boundaries.
\end{abstract}

\section{Introduction} \label{sec:introduction}

Lying at the intersection of information security and natural language processing, linguistic steganography is the practice of hiding information in cover texts~\citep{simmons1984prisoners,anderson1998limits,Bennett2004}.
Formally, the sender \textit{Alice} encodes a secret message, usually in the form of a bit sequence, into a cover text, while the receiver \textit{Bob} decodes the message.
The most important requirement is \textit{security}: The cover text must be so natural that even if transmitted in a public channel, it does not arouse the suspicion of the eavesdropper \textit{Eve}.
In fact, steganography engages in an arms race with \textit{steganalysis}, the practice of detecting the presence of secret messages~\citep{fridrich2009steganography}.
With the security requirement fulfilled, we also want to increase \textit{payload capacity}, the size of the secret message relative to the size of the cover text~\citep{chang-clark-2014-practical}.

Compared with dominant cover media in steganography, such as images, videos, and audio~\citep{fridrich2009steganography}, texts are characterized by a low degree of redundancy.
This makes it particularly challenging to enumerate natural variations of text into which bit chunks are encoded~\citep{chang-clark-2014-practical}.
Nevertheless, this difficulty is surmounted to some degree by powerful neural language models (LMs) for their ability to suggest probable next tokens in a context-aware manner~\citep{fang-etal-2017-generating}, and the research focus has shifted towards increasing payload capacity~\citep{dai-cai-2019-towards,ziegler-etal-2019-neural,shen-etal-2020-near,zhang-etal-2021-provably}.

\begin{figure*}[t]
\centering
\includegraphics[width=\linewidth]{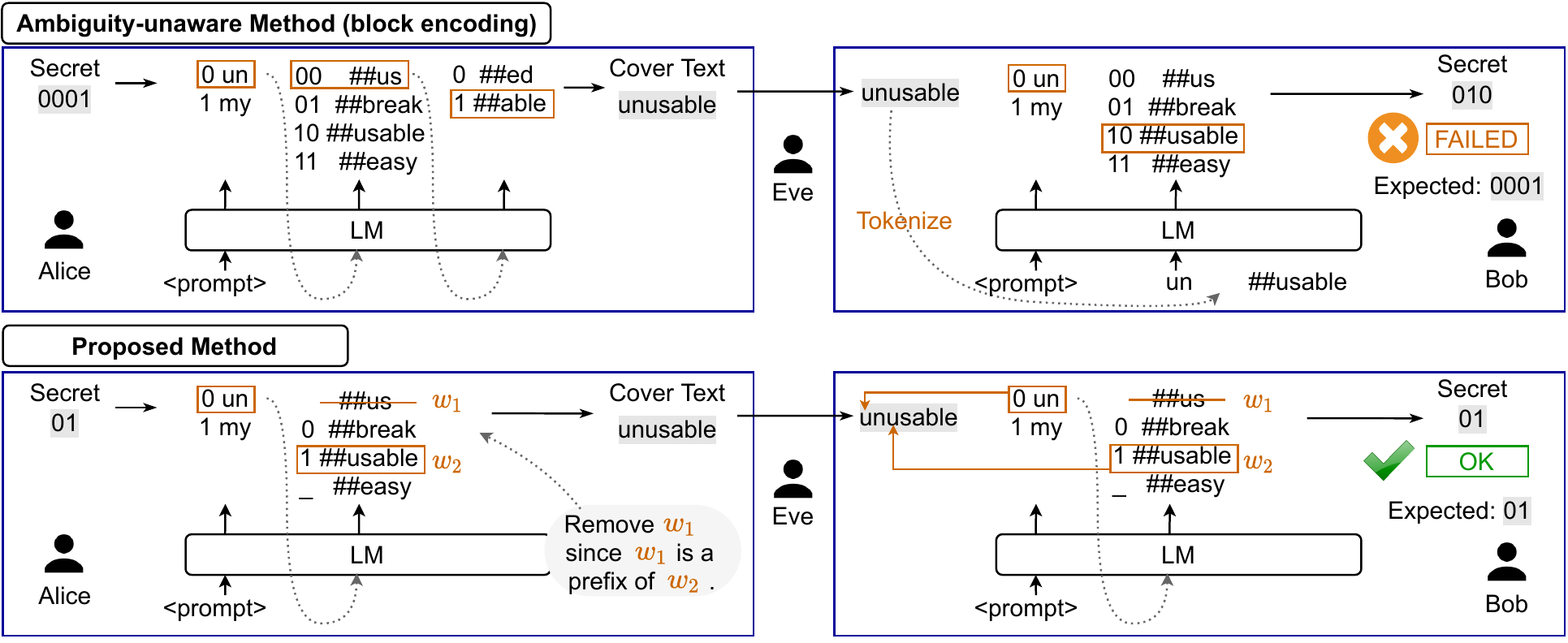}
\caption{Overview of neural linguistic steganography based on an ambiguity-unaware method~(top) and the proposed method~(bottom).
Starting with some introductory context (prompt), the sender Alice iteratively uses a language model (LM) to propose probable next tokens, assigns bit chunks to them, and selects a token corresponding to the secret message. 
The receiver Bob tries to decode the secret message but may fail with the ambiguity-unaware method because the original tokens are not always recovered from the detokenized cover text.
The proposed method guarantees correct decoding by performing stepwise tokenization at Bob's side and by resolving ambiguities.
}
\label{fig:proposed}
\end{figure*}

Previous studies, however, overlook the fact that Alice must detokenize texts before sending them to a public channel; Otherwise they arouse Eve's suspicion.
\citet{ueoka-etal-2021-frustratingly} were the first to point out that Bob may fail to recover the original tokens from detokenized texts, leading to decoding failures.
While segmentation ambiguity has been a vexing problem for \textit{scriptio continua}, or writing systems without explicit word boundaries (e.g., Chinese and Japanese), the near-ubiquitous use of subwords implies that it now affects any language.
For example, suppose that Alice generates the English sequence ``\textit{un} \textit{\#\#us} \textit{\#\#able}''.
Detokenized into ``\textit{unusable}'', it is unfortunately re-tokenized into ``\textit{un} \textit{\#\#usable}'' by Bob (Figure~\ref{fig:proposed} (top)).

While recent proposals are flawed, the fact that the problem went unnoticed till \citet{ueoka-etal-2021-frustratingly} suggests that the errors occur only infrequently.
This leads us to the following question: How often do decoding failures occur?
We expect that they affect morphologically rich languages and \textit{scriptio continua} more severely than English.
We report our experimental results using Russian and Japanese in addition to English.

Although \citet{ueoka-etal-2021-frustratingly} proposed a simple solution for their edit-based method, it is not applicable to LM-based (generation-based) methods.
This motivates us to address the second question: How can generation-based methods overcome segmentation ambiguity?

In this paper, we propose a combination of simple tricks to ensure that Bob recovers the same tokens as Alice (Figure~\ref{fig:proposed} (bottom)).
The proposed method can be applied not only to subword-based LMs but also to \textit{scriptio continua}, as we demonstrate for Japanese.
Our code is available at \url{https://github.com/jumon/himitsu}.

\section{Related Work} \label{sec:related}
\subsection{Finite Word-level Vocabularies}
\label{sec:related-lm}
Before the widespread adoption of subwords, which coincided with the invention of the Transformer architecture~\citep{Vaswani:NIPS2017}, recurrent neural network-based (RNN-based) LMs were accompanied by a finite word-level vocabulary~\citep{Bengio:JMLR2003}.
Vocabulary selection was usually based on frequencies in the training data, and low-frequency words were replaced with the special token \texttt{UNK}.
Applying this technique to linguistic steganography~\citep{zhang-etal-2021-provably} is impractical because \texttt{UNK} is a clear signal of automatic generation and hence is subject to steganalysis.

Oddly enough, previous studies exploring RNN LMs for linguistic steganography~\citep{fang-etal-2017-generating,YangCCS2018,Yang:TIFS2019,Yang:DFW2020,Kang:EI2020,Yang:TIFS2021,Li:TDS2021,Zhou:TDSC2021} make no mention of or obscure the vocabulary selection step.
At any rate, a finite word-level vocabulary should be seen as a security vulnerability.
The complete absence of rare words can be exploited by steganalysis.

\subsection{Subwords in Linguistic Steganography} \label{sec:related-subwords}

In their experiments, \citet{dai-cai-2019-towards}, \citet{ziegler-etal-2019-neural}, and \citet{shen-etal-2020-near} built their steganographic models on top of GPT-2~\citep{radford2019language}, which used subwords.
\citet{dai-cai-2019-towards} and \citet{shen-etal-2020-near} make explicit claims about the applicability of their methods to subword-level LMs.
As we discussed in Section~\ref{sec:introduction}, however, they do not guarantee 100\% recovery of the original subword tokens at Bob's side if Alice detokenizes subwords in order not to arouse Eve's suspicion.

\citet{ueoka-etal-2021-frustratingly} point out that segmentation ambiguity may lead to decoding failures in linguistic steganography.
Their solution is to simply skip subwords.
This is possible because they edit human-generated texts by masking a small portion of tokens~\citep{devlin-etal-2019-bert}, meaning that the resultant texts still contain rare words as before.
If a similar technique is applied to a generation-based method, it falls back into the same problem as LMs with finite word-level vocabularies: the complete absence of rare words.
Note that \citet{ueoka-etal-2021-frustratingly} do not overcome segmentation ambiguity stemming from \textit{scriptio continua} as we do for generation-based steganography in this paper.

Unfortunately, publications that postdate \citet{ueoka-etal-2021-frustratingly} remain silent on segmentation ambiguity.
\citet{Yang:arXiv2202.03795} do not detokenize cover texts at all.
\citet{Yi:SPL2022}, \citet{zheng2022autoregressive}, and \citet{Cao:TCSS2022} make no single mention of subwords even though they used subword-baed models in their experiments.
A faithful
implementation of their methods would lead to decoding failures if detokenization is applied.
For example, \citet{Yi:SPL2022} generate a cover text by interleaving a text-based secret message with dummy words.
While Bob is supposed to be informed of word positions of a secret message in the cover text, subwords do distort word-level positions.

We urge the community to take detokenization and retokenization as necessary steps for linguistic steganography.
Clarification on the use of subwords is also needed.

\section{Segmentation Ambiguity} \label{sec:segmentation-ambiguity}
The basic idea underlying generation-based neural linguistic steganography is to let a powerful neural LM, like GPT-2, enumerate natural variations of text into which bit chunks are encoded~(Figure~\ref{fig:proposed}~(top)).
We assume that Alice and Bob share the LM and an encoding strategy in advance.
Following \citet{ziegler-etal-2019-neural}, we also assume that Alice uses some introductory context (prompt) in a way such that Bob can use the same prompt during decoding.
This helps diversify cover texts.

Now we consider an \textit{ambiguity-unaware} method of generation-based steganography.
For simplicity, we use block encoding~\citep{fang-etal-2017-generating} as the encoding strategy.
At Alice's side, the LM is given a prompt and proposes probable next tokens at each time step.
Alice sorts tokens in descending order of probability and performs a two-step filtering to select the top $2^n$ tokens.
She first selects $c$ tokens with probabilities greater than or equal to $p$ and then chooses $n$ such that it is the largest integer that satisfies $2^n \leq c$.
Each of the tokens is given a unique bit chunk of length $n$, and Alice chooses the one that corresponds to the next $n$ bits of the secret message.
Alice repeats this until she finishes encoding the message.
In the end, she detokenizes the text and sends it to Bob via a public channel.

Receiving the cover text, Bob first tokenizes it and then feeds the resultant tokens to the LM.
He associates tokens with bit chunks in the same way as Alice.
He decodes the secret message by repeatedly selecting a bit chunk corresponding to the next input token.

Unfortunately, this method is flawed because detokenization triggers segmentation ambiguity.
Even if Alice generates the tokens ``\textit{un} \textit{\#\#us} \textit{\#\#able}'', Bob obtains ``\textit{un} \textit{\#\#usable}'', which results in a wrong secret message.
One might be tempted to use an error correcting code for the secret message, but it is of little help because one segmentation error affects all subsequent tokens.

\section{Proposed Method} \label{sec:proposed}
Figure~\ref{fig:proposed} (bottom) shows an overview of the proposed method.
To overcome the segmentation ambiguity problem in generation-based neural linguistic steganography, we combine two simple tricks: \textit{stepwise tokenization} and \textit{token disambiguation}.

\paragraph{Stepwise tokenization}
The first trick is to resist the temptation to use an off-the-shelf tokenizer at Bob's side.
Bob is to imitate Alice's autoregressive generation process instead.
At each time step, Bob selects a token that is a prefix of the remaining part of the detokenized cover text.
For example, suppose that Bob receives the cover text ``\textit{unusable}''.
He first selects ``\textit{un}'', which is a prefix of ``\textit{unusable}''.
Given the remaining part of the cover text, ``\textit{\#\#usable}'', he next selects a prefix of it. 
He repeats this until he finishes reading the cover text.

\paragraph{Token disambiguation}
Stepwise tokenization alone does not resolve segmentation ambiguity.
At the second step of the aforementioned example, Bob faces an indeterminacy problem, as both ``\textit{\#\#us}'' and ``\textit{\#\#usable}'' are prefixes of ``\textit{\#\#usable}''.
We resolve ambiguity by introducing a simple trick at the filtering step of both sides:
If there are two candidate tokens $w_1$ and $w_2$ such that $w_1$ is a prefix of $w_2$, $w_1$ is removed from the candidate list.
For the example above, Alice drops ``\textit{\#\#us}'' because it is a prefix of another candidate ``\textit{\#\#usable}''.
Bob follows the same procedure as Alice to ensure that he can uniquely and correctly identify tokens.

\begin{table*}[t]
\centering
\begin{tabular}{l|cc|cc|cc}
\hline
& \multicolumn{2}{c}{Japanese} & \multicolumn{2}{c}{Russian} & \multicolumn{2}{c}{English}  \\
& Error Rate & Bits/Token & Error Rate & Bits/Token & Error Rate & Bits/Token \\
Method & (\%)$\downarrow$ & $\uparrow$ & (\%)$\downarrow$ & $\uparrow$ & (\%)$\downarrow$ & $\uparrow$ \\
\hline
Ambiguity-unaware & 6.25 & \textbf{2.47} & 3.89 & \textbf{2.52} & 1.18 & \textbf{2.70}  \\
\textbf{Proposed} & \textbf{0.00} & 2.28 & \textbf{0.00} & 2.41 & \textbf{0.00} & 2.59 \\
\hline
\end{tabular}
\caption{Decoding error rates and payload capacity (bits/token) in three different languages.}
\label{tab:main}
\end{table*}

\begin{table}[t]
\centering
\begin{tabular}{l|l}
\hline
\multicolumn{2}{l}{Japanese} \\
\hline
Alice & ... | \begin{CJK}{UTF8}{ipxm}を\end{CJK} | \begin{CJK}{UTF8}{ipxm}成功\end{CJK} | \begin{CJK}{UTF8}{ipxm}させる\end{CJK} | ... \\
Bob & ... | \begin{CJK}{UTF8}{ipxm}を成功させ\end{CJK} | \begin{CJK}{UTF8}{ipxm}る\end{CJK} | ... \\
\hline
\hline
\multicolumn{2}{l}{Russian} \\
\hline
Alice & ... |\selectlanguage{russian} переи \selectlanguage{english}| \#\#\selectlanguage{russian}да \selectlanguage{english}| \#\#\selectlanguage{russian}валось \selectlanguage{english}| ... \\
Bob & ... |\selectlanguage{russian} переи \selectlanguage{english}| \#\#\selectlanguage{russian}дав \selectlanguage{english}| \#\#\selectlanguage{russian}алось \selectlanguage{english}| ... \\
\hline
\hline
\multicolumn{2}{l}{English} \\
\hline
Alice & ... | med | \#\#iation | ... \\
Bob & ... | mediation | ... \\
\hline
\end{tabular}
\caption{Examples of cover texts for which the ambiguity-unaware method caused decoding failures.
A vertical bar marks a token boundary.}
\label{tab:ambiguity}
\end{table}

\section{Experiments} \label{sec:experiments}
We compared the proposed method with the above-mentioned ambiguity-unaware method.
For each method, we generated 10,000 cover texts following different prompts.
Our primary focus was on decoding error rates, or the percentages of decoding failures among the 10,000 trials.
A trial was deemed a failure if Bob re-tokenized the cover text differently from Alice.
The proposed method is guaranteed to have a 0\% decoding error rate, and we intended to experimentally confirm this.
We also evaluated these methods in terms of payload capacity and security.

\subsection{Datasets and Models}

\paragraph{Datasets}
We chose three languages, Japanese, Russian, and English, for which GPT-2 models were available.
For each language, 10,000 lines of text were extracted from the CC-100 web corpus~\citep{wenzek-etal-2020-ccnet} and used as prompts of the LM.
The length of a prompt was 30 characters for Japanese and 10 words for Russian and English.
We used 64 random bits as a secret message.

\paragraph{Models}
We used medium-sized GPT-2 models taken from Hugging Face’s \texttt{transformers} package\footnote{Publicly available at \url{https://huggingface.co/} (Japanese: \href{https://huggingface.co/rinna/japanese-gpt2-medium}{rinna/japanese-gpt2-medium}, Russian: \href{https://huggingface.co/sberbank-ai/rugpt3medium\_based\_on\_gpt2}{sberbank-ai/rugpt3medium\_based\_on\_gpt2}, and English: \href{https://huggingface.co/gpt2-medium}{gpt2-medium}).
Each model had about 350M parameters.}~\citep{wolf-etal-2020-transformers}.
While the Japanese model used SentencePiece~\citep{kudo-richardson-2018-sentencepiece} for its vocabulary, the Russian and English models used a byte-level version of BPE~\citep{radford2019language}.
Accordingly, the prefixes in the proposed method were determined at the byte level.
The probability threshold, $p$, was set to $0.01$.

\begin{table}[t]
\centering
\begin{tabular}{l|c|c|c}
\hline
& \multicolumn{3}{c}{Accuracy (\%)$\downarrow$} \\
Method & ja & ru & en \\
\hline
Ambiguity-unaware & \textbf{86.6} & \textbf{85.4} & \textbf{88.2}  \\
\textbf{Proposed} & 88.6 & 86.5 & 91.5 \\
\hline
(GPT-2 Random) & 79.0 & 77.8 & 82.8 \\
\hline
\end{tabular}
\caption{Results of automatic detection.
The last row shows a baseline that did not encode any secret message.
}
\label{tab:automatic-detection}
\end{table}

\subsection{Automatic Detection (Steganalysis)}
To measure the security of each method, we trained a discriminator to distinguish real texts from generated texts and evaluated the detection accuracy (the lower, the better).
Specifically, we fine-tuned a BERT model on the binary classification task.
As a simple baseline, we also evaluated texts randomly generated by GPT-2, without encoding any secret message.
See Appendix~\ref{sec:appendix-a}
for details.

\subsection{Results}
Table~\ref{tab:main} compares the two methods in terms of decoding error rate and payload capacity.
The error rates for the ambiguity-unaware method were small but non-negligible.
Note that in real situations, secret messages can be longer than 64 bits and consequently can push the decoding error rate upward.
While not strictly comparable because of differences in hyperparameters and datasets, the three languages exhibit an interesting inclination: Japanese, the language without explicit word boundary markers, was the most susceptible to segmentation ambiguity, which was followed firstly by morphologically rich Russian and lastly by analytic English.
Some examples of segmentation ambiguity of the ambiguity-unaware method are shown in Table~\ref{tab:ambiguity} (see Appendix~\ref{sec:appendix-b}
for more examples).

The proposed method featured 100\% correct decoding.
It was at the expense of payload capacity, but no language showed more than a 10\% drop.

Table~\ref{tab:automatic-detection} shows the result of automatic detection.
The proposed method was slightly more prone to automatic detection than the ambiguity-unaware method.
We suspect that the token disambiguation trick worsened the statistical deviation from human-written texts.
The drop in performance is, however, not a prime cause of concern given that even the GPT-2 random baseline was easily detected.
Switching to a more powerful LM would mitigate the risk.
Finally, Appendix~\ref{sec:appendix-c} shows some examples of generated texts.

\section{Discussion} \label{sec:discussion}

Although recent studies on generation-based neural linguistic steganography~\citep{dai-cai-2019-towards,ziegler-etal-2019-neural,shen-etal-2020-near,zhang-etal-2021-provably} exploit the entire vocabulary distributions proposed by an LM, we turn back to na\"{i}ve block encoding~\citep{fang-etal-2017-generating}, which only uses the most probable $2^n$ tokens.
In fact, our solution in its current form is not compatible with the use of the entire vocabulary because with $p=0$, the token disambiguation trick always drops a fixed portion of the vocabulary.
The present study should be seen as a proof-of-concept demonstration focusing on segmentation ambiguity.
We hope that it sets out a future research direction.

\section{Conclusions} \label{sec:conclusions}
Linguistic steganography is an interdisciplinary research area that combines information security and natural language processing (NLP).
In this paper, we investigated its unexpected connection to the decades-old NLP task of word segmentation.
Specifically, we shed light on segmentation ambiguity in generation-based neural linguistic steganography.
Previously proposed methods are flawed if combined with a subword-level LM.

We proposed a combination of simple tricks to guarantee the recovery of the original tokens and thus the correct decoding of a secret message.
Our solution is language-agnostic and is applicable even if no word boundaries are marked.

With powerful neural LMs, linguistic steganography is approaching the level of practical utility.
Now is the time to face up to the fact that without detokenization, linguistic steganography is useless.

\section*{Ethical Considerations}
Linguistic steganography conceals a secret message into a text, without a sign that secret communication is taking place.
With the advance in neural language models, it is becoming possible to generate more natural texts while encoding a good amount of secret data.
The proposed method is language-agnostic and guarantees the correct decoding of a secret message, thus making a step toward real-life applications.
Intended applications of steganography are embedding copyright information, countering censorship, and just for fun, among others.
However, it can also be used to transfer malicious contents, which makes steganography a dual-use technology.
Therefore, along with steganography, steganalysis, the study of detecting the presence of hidden messages, would also be an encouraging research direction to safeguard against malicious use.

\bibliography{anthology,custom}
\bibliographystyle{acl_natbib}

\newpage
\appendix
\setcounter{table}{0}
\renewcommand{\thetable}{A.\arabic{table}}

\section{Details of Automatic Detection} \label{sec:appendix-a}
The 10,000 texts generated by each method were split in an 8:1:1 ratio to create the training, development, and test sets.
For the GPT-2 random baseline, we fed the same prompts to GPT-2 and performed random sampling according to the probabilities of the next tokens.
The real texts and the texts generated by the GPT-2 random baseline were truncated so that they had comparable lengths with texts generated by steganographic methods.
As a discriminator for each language, we used a \texttt{base}-sized BERT model taken from Hugging Face’s \texttt{transformers} package (Japanese: \href{https://huggingface.co/cl-tohoku/bert-base-japanese-whole-word-masking}{cl-tohoku/bert-base-japanese-whole-word-masking}, Russian: \href{https://huggingface.co/DeepPavlov/rubert-base-cased}{DeepPavlov/rubert-base-cased}, and English: \href{https://huggingface.co/bert-base-cased}{bert-base-cased}).
The numbers of parameters of the Japanese, Russian, and English BERT models were about 111M, 178M, and 108M, respectively.

To fine-tune a BERT model, we used generated texts following the prompts as inputs.
Adam~\citep{kingma2015adam} was used as the optimizer with a learning rate of $10^{-5}$.
The batch size was set to 32.
We did not conduct any hyperparameter search and we report the experimental results of single runs.
We trained each model for 10 epochs and used the checkpoint with the best validation accuracy as the final model.
Throughout training, we used a single Quadro P6000 GPU. It took about 15 minutes to train a model.

\section{Examples of segmentation ambiguity} \label{sec:appendix-b}
Table~\ref{tab:appendix-b} shows more examples of cover texts for which the ambiguity-unaware method caused decoding failures.

\begin{table*}[t]
\centering
\begin{tabular}{l|l}
\hline
\multicolumn{2}{l}{Japanese} \\
\hline
Alice & ... | \begin{CJK}{UTF8}{ipxm}新しい\end{CJK} | \begin{CJK}{UTF8}{ipxm}カ\end{CJK} | \begin{CJK}{UTF8}{ipxm}ラム\end{CJK} | \begin{CJK}{UTF8}{ipxm}\textcolor{red}{を}\end{CJK} | \begin{CJK}{UTF8}{ipxm}\textcolor{red}{作成した}\end{CJK} | \begin{CJK}{UTF8}{ipxm}\textcolor{red}{ら}\end{CJK} | \begin{CJK}{UTF8}{ipxm}どう\end{CJK} | \begin{CJK}{UTF8}{ipxm}します\end{CJK} | \begin{CJK}{UTF8}{ipxm}か\end{CJK} | \begin{CJK}{UTF8}{ipxm}?\end{CJK} | ... \\
Bob & ... | \begin{CJK}{UTF8}{ipxm}新しい\end{CJK} | \begin{CJK}{UTF8}{ipxm}カ\end{CJK} | \begin{CJK}{UTF8}{ipxm}ラム\end{CJK} | \begin{CJK}{UTF8}{ipxm}\textcolor{red}{を作成し}\end{CJK} | \begin{CJK}{UTF8}{ipxm}\textcolor{red}{たら}\end{CJK} | \begin{CJK}{UTF8}{ipxm}どう\end{CJK} | \begin{CJK}{UTF8}{ipxm}します\end{CJK} | \begin{CJK}{UTF8}{ipxm}か\end{CJK} | \begin{CJK}{UTF8}{ipxm}?\end{CJK} | ... \\
\hline
Alice & ... |  \begin{CJK}{UTF8}{ipxm}各\end{CJK} | \begin{CJK}{UTF8}{ipxm}会場\end{CJK} | \begin{CJK}{UTF8}{ipxm}で\end{CJK} | \begin{CJK}{UTF8}{ipxm}撮影した\end{CJK} | \begin{CJK}{UTF8}{ipxm}写真を\end{CJK} | \begin{CJK}{UTF8}{ipxm}1\end{CJK} | \begin{CJK}{UTF8}{ipxm}\textcolor{red}{枚の}\end{CJK} | \begin{CJK}{UTF8}{ipxm}\textcolor{red}{アルバム}\end{CJK} | \begin{CJK}{UTF8}{ipxm}にして\end{CJK} | \begin{CJK}{UTF8}{ipxm}配布\end{CJK} | \begin{CJK}{UTF8}{ipxm}される\end{CJK} | ... \\
Bob & ... |  \begin{CJK}{UTF8}{ipxm}各\end{CJK} | \begin{CJK}{UTF8}{ipxm}会場\end{CJK} | \begin{CJK}{UTF8}{ipxm}で\end{CJK} | \begin{CJK}{UTF8}{ipxm}撮影した\end{CJK} | \begin{CJK}{UTF8}{ipxm}写真を\end{CJK} | \begin{CJK}{UTF8}{ipxm}1\end{CJK} | \begin{CJK}{UTF8}{ipxm}\textcolor{red}{枚のアルバム}\end{CJK} | \begin{CJK}{UTF8}{ipxm}にして\end{CJK} | \begin{CJK}{UTF8}{ipxm}配布\end{CJK} | \begin{CJK}{UTF8}{ipxm}される\end{CJK} | ... \\
\hline
\hline
\multicolumn{2}{l}{Russian} \\
\hline
Alice & ... |\selectlanguage{russian} Он  \selectlanguage{english}|\selectlanguage{russian} достаточно  \selectlanguage{english}|\selectlanguage{russian} лак  \selectlanguage{english}|\selectlanguage{russian} \textcolor{red}{\#\#о}  \selectlanguage{english}|\selectlanguage{russian} \textcolor{red}{\#\#нич}  \selectlanguage{english}|\selectlanguage{russian} \textcolor{red}{\#\#ен}  \selectlanguage{english}|\selectlanguage{russian} и  \selectlanguage{english}|\selectlanguage{russian} в  \selectlanguage{english}|\selectlanguage{russian} тоже  \selectlanguage{english}|\selectlanguage{russian} время  \selectlanguage{english}|\selectlanguage{russian} очень  \selectlanguage{english}| ... \\
Bob & ... |\selectlanguage{russian} Он  \selectlanguage{english}|\selectlanguage{russian} достаточно  \selectlanguage{english}|\selectlanguage{russian} лак  \selectlanguage{english}|\selectlanguage{russian} \textcolor{red}{\#\#они}  \selectlanguage{english}|\selectlanguage{russian} \textcolor{red}{\#\#чен}  \selectlanguage{english}|\selectlanguage{russian} и  \selectlanguage{english}|\selectlanguage{russian} в  \selectlanguage{english}|\selectlanguage{russian} тоже  \selectlanguage{english}|\selectlanguage{russian} время  \selectlanguage{english}|\selectlanguage{russian} очень  \selectlanguage{english}| ... \\
\hline
Alice & ... |\selectlanguage{russian} при \selectlanguage{english}|\selectlanguage{russian} любом \selectlanguage{english}|\selectlanguage{russian} исходе \selectlanguage{english}|\selectlanguage{russian} дела \selectlanguage{english}|\selectlanguage{russian} он \selectlanguage{english}|\selectlanguage{russian} стремится \selectlanguage{english}|\selectlanguage{russian} получить \selectlanguage{english}|\selectlanguage{russian} \textcolor{red}{максим} \selectlanguage{english}|\selectlanguage{russian} \textcolor{red}{\#\#альную} \selectlanguage{english}| ... \\
Bob & ... |\selectlanguage{russian} при \selectlanguage{english}|\selectlanguage{russian} любом \selectlanguage{english}|\selectlanguage{russian} исходе \selectlanguage{english}|\selectlanguage{russian} дела \selectlanguage{english}|\selectlanguage{russian} он \selectlanguage{english}|\selectlanguage{russian} стремится \selectlanguage{english}|\selectlanguage{russian} получить \selectlanguage{english}|\selectlanguage{russian} \textcolor{red}{максимальную} \selectlanguage{english}| ... \\
\hline
\hline
\multicolumn{2}{l}{English} \\
\hline
Alice & ... |  decreased | more | in | dry | land | environments | compared | to | \textcolor{red}{wet} | \textcolor{red}{\#\#lands} | ... \\
Bob & ... | decreased | more | in | dry | land | environments | compared | to | \textcolor{red}{wetlands} | ... \\
\hline
Alice & ... | while | \textcolor{red}{C} | \textcolor{red}{\#\#CA} | \textcolor{red}{\#\#A} | works | directly | with | local | schools | ... \\
Bob & ... | while | \textcolor{red}{CC} | \textcolor{red}{\#\#AA} | works | directly | with | local | schools | ... \\
\hline
\end{tabular}
\caption{More examples of cover texts for which the ambiguity-unaware method caused decoding failures.
A vertical bar marks a token boundary, and tokens colored red indicate where tokenization differs.}
\label{tab:appendix-b}
\end{table*}

\section{Examples of texts generated by our proposed method} \label{sec:appendix-c}
Table~\ref{tab:appendix-c} shows examples of texts generated by the proposed method following prompts.
Each generated text carries a 64-bit random secret message.

\begin{table*}[t]
\centering
\begin{tabular}{l|l}
\hline
\multicolumn{2}{l}{Japanese} \\
\hline
Prompt & \begin{CJK}{UTF8}{ipxm}お店ではたくさんの商品が所狭しと並んでいます。それだけ置いて\end{CJK} \\
Generated text & \begin{CJK}{UTF8}{ipxm}おいて欲しい商品だからでしょう。 そんな風に商品が置いてあります\end{CJK} \\
& \begin{CJK}{UTF8}{ipxm}ので、商品を購入するにはお目当てのお店に行くことになるのが通常\end{CJK} \\
\hline
Prompt & \begin{CJK}{UTF8}{ipxm}私の場合は、仕事の帰宅時間がとても不規則だったので睡眠時間が\end{CJK} \\
Generated text & \begin{CJK}{UTF8}{ipxm}不規則な状態でしたが、今となってはその理由も分かる気もします(;'∀')」 \end{CJK} \\
& \begin{CJK}{UTF8}{ipxm}これは私も同じ経験なのですが、「おやすみ\end{CJK} \\
\hline
\hline
\multicolumn{2}{l}{Russian} \\
\hline
Prompt & \selectlanguage{russian}Обоих не было в игре, но ответят, как зайдут. Других\selectlanguage{english} \\
Generated text & \selectlanguage{russian}тоже. Но, судя по всему, у вас с собой будет оружие и запасные\selectlanguage{english} \\
& \selectlanguage{russian}патроны к ним. Это я уже от себя\selectlanguage{english} \\
\hline
Prompt & \selectlanguage{russian}Это меня еще больше встревожило. Несколько часов я провел без\selectlanguage{english} \\
Generated text & \selectlanguage{russian}сна. Но ничего. Я еще буду в порядке, когда у нас появятся свои\selectlanguage{english}\\
& \selectlanguage{russian}дома… Но тут зазвонил мобильный\selectlanguage{english} \\
\hline
\hline
\multicolumn{2}{l}{English} \\
\hline
Prompt & She hugged me then, burying her face into my chest. \\
Generated text & It hurt me too much and I was getting hot and sweaty, and I had a terrible stomach \\
&  bug. It didn \\
\hline
Prompt & I have read many articles on the subject and have \\
Generated text & tried not to comment on this as it has become the focus of an intense debate amongst \\
& fans in my time with this \\
\hline
\end{tabular}
\caption{Examples of texts generated by the proposed method following prompts.
Each generated text carries a 64-bit random secret message.
Following \citet{ziegler-etal-2019-neural}, we stop generation when the proposed method finishes embedding the message.
}
\label{tab:appendix-c}
\end{table*}

\end{document}